\documentclass[runningheads]{llncs}
\usepackage[T1]{fontenc}
\usepackage{multirow}
\usepackage{tabularx}
\usepackage[linesnumbered,ruled,vlined]{algorithm2e}
\usepackage[table]{xcolor}
\usepackage{booktabs}
\usepackage{amsmath,amssymb}
\usepackage{pifont} 
\usepackage{graphicx}
\usepackage{colortbl}
\usepackage{lipsum} 
\usepackage{tgheros}
\usepackage{graphicx,verbatim}
\usepackage{threeparttable}

\newcolumntype{Y}{>{\centering\arraybackslash}X}

\begin{document}
\title{Debunking Optimization Myths in Federated Learning for Medical Image Classification}
\titlerunning{Debunking Optimization Myths in FL for Medical Image Classification}
\author{Youngjoon Lee\inst{1}\textsuperscript{*}\and
Hyukjoon Lee\inst{2}\and
Jinu Gong\inst{3}\and\\
Yang Cao\inst{4}\and
Joonhyuk Kang\inst{1}\textsuperscript{\dag}}

\authorrunning{Y. Lee et al.}
%
\institute{School of Electrical Engineering, KAIST, South Korea\and
AI Group, AMD, United States \and
Department of Applied AI, Hansung University, South Korea \and
Department of Computer Science, Institute of Science Tokyo, Japan\\
\email{yjlee22@kaist.ac.kr}, \email{jkang@kaist.ac.kr}
}

\maketitle

\renewcommand{\thefootnote}{*}
\footnotetext{Work done while a visiting student at Science Tokyo}
\renewcommand{\thefootnote}{\dag}
\footnotetext{Corresponding author}

\begin{abstract}
Federated Learning (FL) is a collaborative learning method that enables decentralized model training while preserving data privacy.
Despite its promise in medical imaging, recent FL methods are often sensitive to local factors such as optimizers and learning rates, limiting their robustness in practical deployments.
In this work, we revisit vanilla FL to clarify the impact of edge device configurations, benchmarking recent FL methods on colorectal pathology and blood cell classification task.
We numerically show that the choice of local optimizer and learning rate has a greater effect on performance than the specific FL method.
Moreover, we find that increasing local training epochs can either enhance or impair convergence, depending on the FL method.
These findings indicate that appropriate edge-specific configuration is more crucial than algorithmic complexity for achieving effective FL.

\end{abstract}

\section{Introduction}
\label{sec:intro}
In recent years, medical imaging has undergone significant advancements, enhancing both diagnostic accuracy and the range of clinical applications in various domains of healthcare \cite{abouelmehdi2018big,rajpurkar2022ai}.  
These improvements have largely resulted from the widespread adoption of deep learning techniques, which leverage large-scale, high-quality datasets to train powerful neural networks \cite{bisio2025ai,whang2023data}.  
However, in many real-world medical scenarios, stringent privacy regulations and institutional governance severely restrict the direct sharing of sensitive data across organizations \cite{joshi2022federated,lee2022accelerated,rauniyar2023federated}.  
To address this challenge, Federated Learning (FL) \cite{mcmahan2017communication} has emerged as a promising decentralized learning paradigm that allows collaborative model training without the need to transmit raw data \cite{ding2022federated,li2020federated}.  
By ensuring that data remain localized at the source, FL offers a privacy-preserving solution that aligns well with legal and ethical constraints in medical AI development \cite{antunes2022federated,lee2023fast}.

FL provides several key advantages that make it particularly well-suited for healthcare-related applications \cite{kairouz2021advances}.  
First, FL preserves data sovereignty by enabling institutions to retain complete control over their datasets while contributing to the global model \cite{rieke2020future}. 
Additionally, this method allows for the development of more generalizable models by leveraging diverse datasets \cite{pfitzner2021federated}. 
Furthermore, FL reduces the computational and storage burden on individual institutions by distributing the training process across multiple participants \cite{guan2024federated}.  
Thus, FL emerges as an effective and scalable solution for developing secure, collaborative, and inclusive medical AI systems \cite{nguyen2022federated}.

\begin{figure}[t]
\centering
\includegraphics[width=11.85cm]{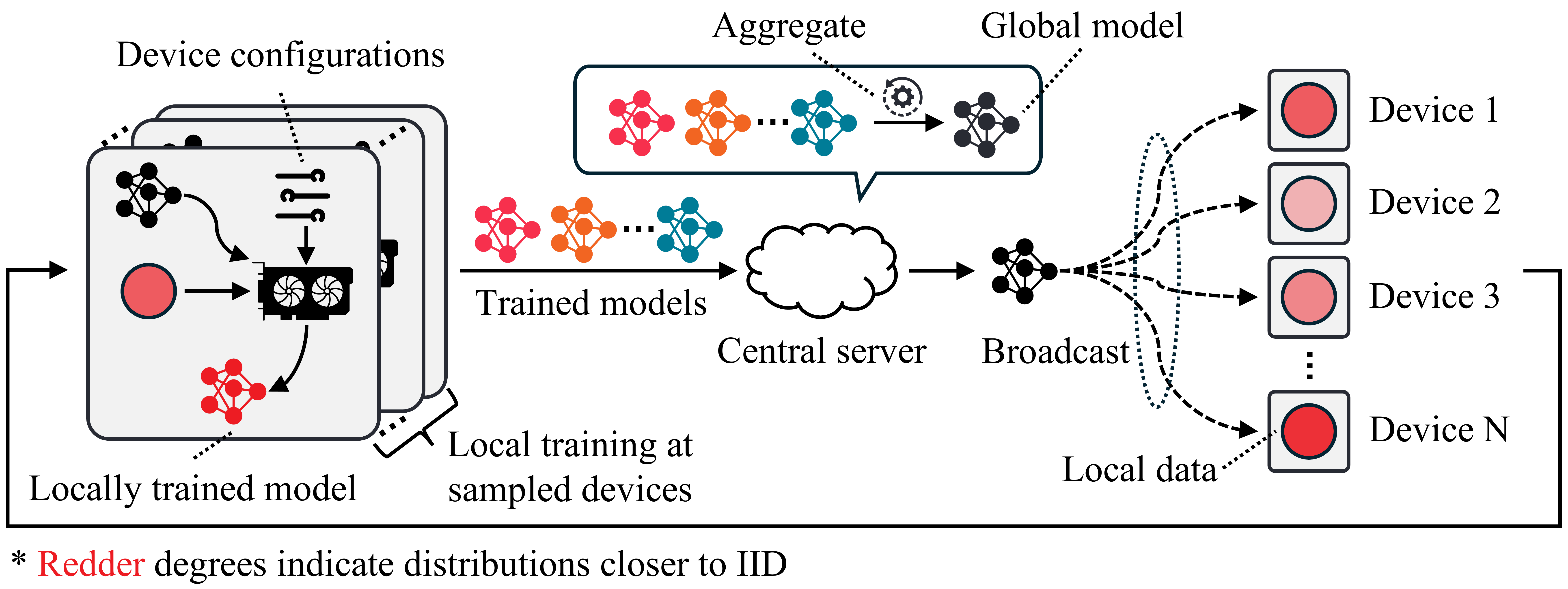}
\caption{Illustration of general FL process. In each global epoch, selected edge devices train locally using their own data and configurations (e.g., local optimizer, local learning rate, local epochs). Updated models are sent to the server for aggregation, and the new global model is broadcast back to all devices.}
\label{fig:intro}
\end{figure}

However, many recent FL methods have introduced complex optimization pipelines that exhibit a strong dependence on the careful tuning of algorithm-specific hyperparameters \cite{lee2025revisit}.  
These methods rely heavily on carefully tuned hyperparameters, regularization factors, and aggregation techniques, sensitive to variations in data distribution and network conditions \cite{houssein2023boosted}.  
Moreover, deploying FL systems in practical environments introduces additional variability through individual edge device configurations.  
In detail, differences in the choice of local optimizers and learning rates can significantly influence convergence behavior and model performance.  

In this work, we revisit the effectiveness of vanilla FL—namely, the original approach based on simple model averaging—within the context of medical imaging.  
We conduct comprehensive experiments on a representative medical imaging dataset using recent FL methods, including vanilla FL. 
Our numerical results demonstrate that vanilla FL performs comparably to more sophisticated methods across a wide range of federated training conditions.  
Despite diverse local optimizers, learning rates, and local epoch settings, relative performance rankings remain stable.
Therefore, we numerically show that the effectiveness of FL in practice depends strongly on the appropriate selection of local training configurations.

\section{Problem and Method}
\label{sec:system}

\subsection{Federated Setting}
We consider two representative data distribution settings in FL, including IID and non-IID.
In the IID setting, each edge device $n \in \{1, \ldots, N\}$ receives local data $\mathcal{D}^n$ sampled independently from a shared distribution $\mathcal{P}$, assumed to be standard normal:
\begin{align}
x \sim \mathcal{P} = \mathcal{N}(0, I), \quad \forall x \in \mathcal{D}^n,\ \forall n.
\end{align}

In contrast, in the non-IID setting, we model label distribution heterogeneity using a Dirichlet distribution as \cite{li2022federated}.  
Let $\mathcal{Y} = \{1, \ldots, C\}$ be the set of class labels.  
For each device $n$, a label distribution vector $\mathbf{p}^n = (p_1^n, \ldots, p_C^n)$ is drawn from a Dirichlet distribution:
\begin{align}
\mathbf{p}^n \sim \text{Dir}(\alpha), \quad \alpha > 0.
\end{align}
Then, the local dataset $\mathcal{D}^n$ is constructed by sampling data conditioned on the assigned label proportions $\mathbf{p}^n$.  
Note that, smaller values of $\alpha$ result in higher label skew, where each device is likely to contain samples from only a few classes. 
This setting reflects real-world data heterogeneity, which induces client drift and thereby hinders effective training in FL.

\subsection{General FL Process}
The goal of FL is to learn a global model $w \in \mathbb{R}^d$ by minimizing the aggregated objective function
$F(w) := \frac{1}{N} \sum_{n=1}^N F_n(w)$.
Each local objective $F_n(w)$ is defined over edge device $n$'s private dataset $\mathcal{D}^n$ as
$F_n(w) := \frac{1}{|\mathcal{D}^n|} \sum_{x \in \mathcal{D}^n} f(w; x)$,
where $f(w; x)$ denotes the loss evaluated on sample $x$.

At the beginning of each global epoch $g_e$, the central server broadcasts the current global model $w^{g_e}$ to all $N$ edge devices.  Then, the server randomly selects a subset of devices $\mathcal{S}_{g_e} \subset \{1, \ldots, N\}$ to participate in federated training during the global epoch.  
Each selected edge device $n \in \mathcal{S}{g_e}$ initializes its local model to $w^{g_e}$ and trains locally using the optimizer $\text{EdgeOpt}(\cdot)$ for $l_e$ steps with local learning rate $\eta_l$:
\begin{align}
w_n^{g_e, l_e} = \text{EdgeOpt}(w^{g_e}, \mathcal{D}^n, \eta_l, l_e).
\end{align}

After completing local training, each participating device sends its updated model $w_n^{g_e, le}$ to the central server.  
The server aggregates these updates using a server-side optimizer $\text{ServerOpt}(\cdot)$, which may incorporate weighting schemes or other update rules:
\begin{align}
w^{g_e+1} = \text{ServerOpt}(\{w_n^{g_e, le}\}_{n \in \mathcal{S}_{g_e}}).
\end{align}

This procedure repeats for $g_e = 1, \ldots, G$, until convergence \cite{li2019convergence} or a predefined number of global epochs $G$ is reached.  
The dynamics of federated training depend on device-specific configurations, including local optimizer, $\eta_l$, and $l_e$, as well as the optimization behavior defined by $\text{EdgeOpt}(\cdot)$ and $\text{ServerOpt}(\cdot)$.

\section{Experiment and Results}
\label{sec:experiment}

\begin{table}[t]
\centering
\caption{Top-1 test accuracy (\%) and corresponding sub-optimal $g_e$ of FL methods under IID and non-IID settings. The vanilla FL achieves performance comparable to recent FL methods. Here, peak-$g_e$ denotes the global epoch at which each method reaches its peak accuracy.}
\label{tab:1}
\scriptsize
\begin{tabularx}{\textwidth}{l*{8}{>{\centering\arraybackslash}X}}
\toprule
\multirow{3}{*}{\textbf{Method}} & \multicolumn{4}{c}{\textbf{Colorectal Pathology Task}} & \multicolumn{4}{c}{\textbf{Blood Cell Task}} \\
\cmidrule(lr){2-5} \cmidrule(lr){6-9}
& \multicolumn{2}{c}{\textbf{IID}} & \multicolumn{2}{c}{\textbf{non-IID}} & \multicolumn{2}{c}{\textbf{IID}} & \multicolumn{2}{c}{\textbf{non-IID}} \\
\cmidrule(lr){2-3} \cmidrule(lr){4-5} \cmidrule(lr){6-7} \cmidrule(lr){8-9}
& \textbf{Acc. (\%)} & \textbf{Peak-$g_e$} & \textbf{Acc. (\%)} & \textbf{Peak-$g_e$} & \textbf{Acc. (\%)} & \textbf{Peak-$g_e$} & \textbf{Acc. (\%)} & \textbf{Peak-$g_e$} \\
\midrule
FedAvg & 95.02 & 52.7 & 93.88 & 90.3 & 96.91 & 94.3 & 92.69 & 95.7 \\
FedDyn & 95.17 & 30.7 & 94.58 & 79.3 & 97.82 & 77.7 & 96.89 & 94.7 \\
FedSAM & 95.21 & 65.7 & 94.09 & 83.0 & 96.99 & 97.7 & 93.57 & 94.0 \\
FedSpeed & 95.38 & 36.0 & 94.73 & 81.7 & 97.94 & 88.0 & 97.16 & 87.0 \\
FedSMOO & 95.37 & 70.3 & 94.81 & 73.0 & 97.99 & 87.0 & 97.15 & 91.7 \\
FedGamma & 93.66 & 71.3 & 89.79 & 70.7 & 96.65 & 98.0 & 93.49 & 99.3 \\

\bottomrule
\end{tabularx}
\end{table}

\subsection{Experiment Setting}
To check the robustness of FedAvg, we conduct experiments on a colorectal pathology \cite{kather2019predicting} and blood cell \cite{acevedo2020dataset} image classification task under federated settings. 
In detail, all edge devices employ ConvNeXtV2 \cite{woo2023convnext} as local AI model, and we compare FedAvg with recent FL methods including FedDyn \cite{acarfederated}, FedSAM \cite{qu2022generalized}, FedSpeed \cite{sunfedspeed}, FedSMOO \cite{sun2023dynamic}, and FedGamma \cite{10269141}.
To simulate the non-IID setting, we distribute samples across $N=100$ edge devices with $\alpha=0.1$.  
At each $g_e$, a subset of $M=10$ devices is randomly selected to participate in training.  
All experiments are run with 3 random seeds, and training is accelerated using AMD Instinct MI300X GPUs \cite{smith2024amd}, supported by AMD Developer Cloud credits.

\subsubsection{FL Method Comparison: IID vs. non-IID}
To investigate whether FedAvg can achieve competitive performance compared to recent FL methods, we evaluate under both IID and non-IID label distributions.
As shown in Table~\ref{tab:1}, all methods attain similar top-1 test accuracy, with only marginal differences across settings.
Under the IID setting, FedAvg achieves 95.02\% and 96.91\% accuracy on the Colorectal Pathology and Blood Cell tasks, respectively—only 0.36\% and 1.03\% below the best-performing methods (FedSpeed and FedSMOO).
In terms of convergence, FedAvg requires 52.7 and 94.3 global epochs, whereas FedDyn and FedSpeed converge fastest with 30.7 and 77.7 epochs, respectively.

Under the non-IID setting, FedAvg achieves 93.88\% on Colorectal and 92.69\% on Blood Cell, falling 0.93\% and 4.47\% short, with FedSMOO and FedSpeed being the top performers for each task.
Moreover, FedAvg shows strong stability: on the Colorectal task, it outperforms FedGamma by 4.09\%, and on the Blood Cell task, it surpasses FedGamma by 3.20\%, while also converging up to 25 epochs earlier.
Overall, the results indicate that the choice of FL algorithm has only a limited effect on the performance.

\begin{figure}[t]
\centering
\includegraphics[width=\columnwidth]{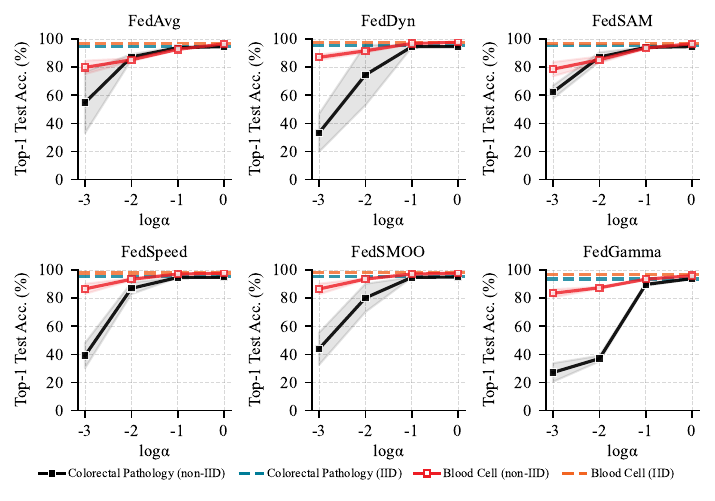}
\caption{Top-1 test accuracy (\%) of FL methods under varying Dirichlet $\alpha$ values, which control the degree of label heterogeneity. While all methods improve as $\alpha$ increases, their robustness at low $\alpha$ varies significantly.}
\label{fig:result4}
\end{figure}

\subsubsection{Impact of non-IID Degree}
To analyze the impact of label distribution skew on FL performance, we adjust the Dirichlet $\alpha$ parameter to simulate varying degrees of non-IID conditions. 
When $\alpha$ becomes larger and label distributions are more balanced, results in consistent performance gains across all FL methods, as shown in Fig. ~\ref{fig:result4}.
This trend persists across both medical imaging tasks, indicating that the degree of data heterogeneity strongly influences model performance. 
Although some methods exhibit slightly better robustness than others, the overall shift in performance is largely driven by the change in $\alpha$. 

In the colorectal task, FedAvg improves dramatically from 54.87\% at $\alpha{=}0.001$ to 94.65\% at $\alpha{=}1.0$, closely tracking FedSAM, which rises from 62.29\% to 94.66\%. 
Even methods like FedGamma, which struggle under high heterogeneity (27.12\% at $\alpha{=}0.001$), recover to over 93\% as $\alpha$ increases. 
For the blood cell task, the best-performing methods—FedSpeed and FedSMOO—reach above 97.7\% at $\alpha{=}1.0$, improving from around 86.6\% at $\alpha{=}0.001$. 
Meanwhile, FedDyn shows a narrower gain, rising from 87.02\% to 97.70\%, suggesting limited responsiveness to distribution shift. 
FedAvg, although simpler, steadily narrows the gap as heterogeneity decreases. 
Collectively, these results imply that adapting to non-IID severity is often more consequential than the specific FL algorithm employed.

\begin{figure}[t]
\centering
\includegraphics[width=\columnwidth]{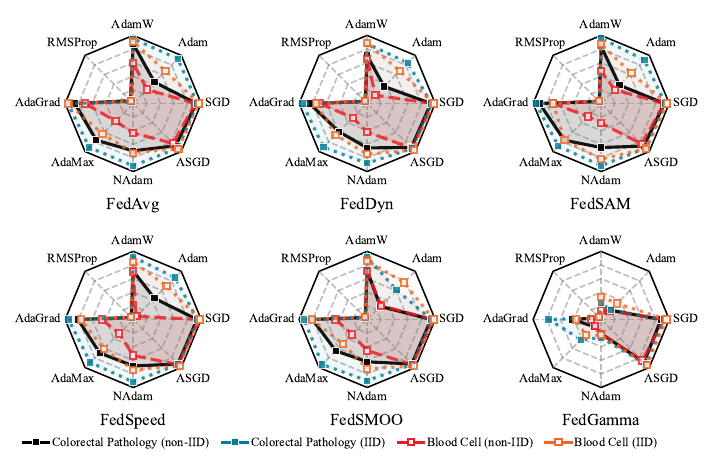}
\caption{Top-1 test accuracy (\%) of FL methods evaluated under representative local optimizers at the edge device. The choice of local optimizer substantially impacts FL performance more than optimization variations.}
\label{fig:result2}
\end{figure}

\subsubsection{Impact of Local Optimizers}
To examine how local optimizer choice affects federated training performance, we evaluate FL methods across commonly used optimizers.  
As shown in Fig.~\ref{fig:result2}, optimizer choice leads to greater performance variation than the specific FL algorithm itself.  
FedAvg shows relatively steady behavior, but its accuracy still drops from over 93\% with SGD to as low as 20\% with RMSProp.  
In contrast, methods like FedDyn and FedGamma exhibit sharp fluctuations, particularly under Adam and AdamW.  
These results highlight the critical role of optimizer stability in federated training, particularly under heterogeneous data distributions. 

When trained with SGD, all methods achieve strong performance, with FedSpeed and FedSMOO exceeding 94\% in both the colorectal pathology and blood cell tasks.  
However, under Adam, FedDyn drops to 45.10\% in the colorectal task and 29.68\% in the blood cell task, while FedGamma falls below 32\% in both.  
While AdaGrad yields more consistent results—FedAvg achieves 87.85\% and 75.61\% in the two tasks respectively—RMSProp consistently underperforms, producing sub-20\% accuracy across the board.  
Among stable alternatives, ASGD performs well overall, with FedSAM and FedDyn both exceeding 90\% in the colorectal task.  
Overall, the local optimizer selection has a greater impact on performance than the choice of FL method.

\begin{figure}[t]
\centering
\includegraphics[width=\columnwidth]{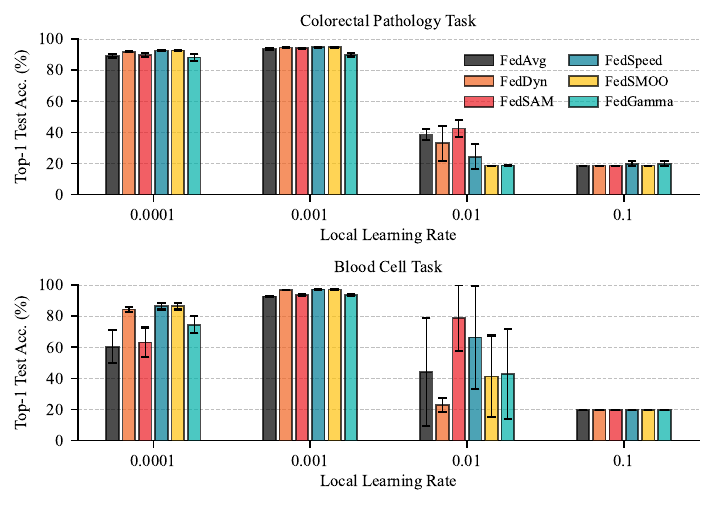}
\caption{Top-1 test accuracy (\%) of FL methods evaluated under various local learning rates. Improper local learning rate selection can significantly degrade FL performance. The error bars denote the standard deviation over 3 independent runs.}
\label{fig:result3}
\end{figure}

\subsubsection{Impact of Local Learning Rate}
To investigate how the local learning rate influences model performance, we evaluate FL methods across three representative values: 0.0001, 0.001, 0.01, and 0.1.  
At $\eta_l = 0.001$, all methods exhibit stable and high accuracy, demonstrating reliable convergence, as shown in Fig.~\ref{fig:result3}.  
However, as $\eta_l$ increases to 0.01, performance variation across methods becomes more pronounced.  
This trend is further exacerbated at 0.1, where most methods collapse to below 20\% accuracy regardless of task.  
These observations indicate that higher learning rates destabilize training, particularly in FL methods sensitive to local updates.

At $\eta_l = 0.001$, FedSMOO and FedSpeed achieve the highest accuracies in the blood cell task, both exceeding 97.1\%, while FedDyn reaches 96.89\%.  
When increasing $\eta_l$ to 0.01, FedSAM remains relatively robust, achieving 78.73\% in the blood cell task and 42.46\% in the colorectal task.  
In contrast, FedSMOO and FedSpeed drop sharply to 18.64\% and 24.37\%, respectively, in the colorectal task.  
This instability becomes universal at $\eta_l = 0.1$, where all methods flatten to around 19.47\% in the blood cell task.  
These results confirm that high learning rates impede convergence, underscoring the importance of carefully tuning  $\eta_l$ to avoid performance degradation.

\begin{table*}[t]
\centering
\scriptsize
\caption{Top-1 test accuracy (\%) difference and corresponding global epoch difference between each FL method and FedAvg across different local epochs. The presented values represent mean differences relative to FedAvg.}
\label{tab:2}
\begin{tabularx}{\textwidth}{l*{12}{>{\centering\arraybackslash}X}}
\toprule
\multirow{2}{*}{\textbf{Method}} & \multicolumn{6}{c}{\textbf{Colorectal Pathology Task}} & \multicolumn{6}{c}{\textbf{Blood Cell Task}} \\
\cmidrule(lr){2-7} \cmidrule(lr){8-13}
& \multicolumn{2}{c}{$l_e=1$} & \multicolumn{2}{c}{$l_e=5$} & \multicolumn{2}{c}{$l_e=20$} & \multicolumn{2}{c}{$l_e=1$} & \multicolumn{2}{c}{$l_e=5$} & \multicolumn{2}{c}{$l_e=20$} \\
\cmidrule(lr){2-3} \cmidrule(lr){4-5} \cmidrule(lr){6-7} \cmidrule(lr){8-9} \cmidrule(lr){10-11} \cmidrule(lr){12-13}
& $\delta_{Acc.}$ & $\delta_{g_e}$ & $\delta_{Acc.}$ & $\delta_{g_e}$ & $\delta_{Acc.}$ & $\delta_{g_e}$ & $\delta_{Acc.}$ & $\delta_{g_e}$ & $\delta_{Acc.}$ & $\delta_{g_e}$ & $\delta_{Acc.}$ & $\delta_{g_e}$ \\
\midrule
FedDyn    & +2.69 & -9.0 & +0.04 & -12.7 & +0.17 & -18.3 & +14.41 & +0.3 & +1.11 & -21.3 & +0.10 & -43.7 \\
FedSAM    & +1.09 & -1.7 & +0.02 & -0.3  & -0.17 & +12.7 & +0.24  & -1.0 & +0.14 & +0.7  & -0.08 & -8.7 \\
FedSpeed  & +3.30 & -4.7 & +0.24 & +5.7  & -0.48 & -16.7 & +13.85 & +0.3 & +1.16 & -6.3  & -0.01 & -28.3 \\
FedSMOO   & +3.52 & +1.0 & +0.43 & +22.7 & -0.39 & -36.3 & +13.87 & +0.3 & +1.18 & -20.3 & +0.09 & -33.3 \\
FedGamma  & +1.20 & -  & -0.70 & +35.0 & -3.46 & -11.0 & +6.03  & -  & -0.53 & +1.0  & -1.72 & -15.7 \\
\bottomrule
\end{tabularx}
\end{table*}

\subsubsection{Ablation Study}
To check the effect of local training duration, we evaluate performance for $l_e \in \{1, 5, 20\}$ using SGD with $\eta_l = 0.001$, as shown in Table \ref{tab:2}.  
In particular, increasing $l_e$ generally improves accuracy for FedAvg, reaching 95.22\% at $l_e=20$, while also reducing required global epochs at $l_e=5$.  
FedDyn benefits from more frequent local updates at $l_e=1$, showing the largest accuracy gains of +2.69\% and +14.41\% in the colorectal and blood cell tasks, respectively.  
In contrast, FedGamma consistently degrades with longer local training, dropping by $-3.46$\% and $-1.72$\% at $l_e=20$ while requiring fewer global epochs.  
FedSMOO and FedSpeed show similar trends: both improve moderately at $l_e=5$ in both tasks, but their accuracy declines at $l_e=20$, despite converging faster than FedAvg.  
In conclusion, longer local training can either benefit or impair performance, highlighting the need for adaptive and well-calibrated $l_e$ settings.

\section{Conclusion}
\label{sec:conclusion}
In this work, we show that FL performance is more sensitive to edge-specific hyperparameters than to the underlying federated optimization strategy.
Through comprehensive experiments, we show that local optimizers and local learning rates impact performance and convergence more than the choice of FL method.
Moreover, the heightened sensitivity of recent FL methods to these hyperparameters raises concerns regarding their robustness, reproducibility, and deployability in practical settings.
These results highlight the need for FL methods that are not only theoretically sound but also resilient to variations in device-level configurations.

\begin{credits}
\subsubsection{\ackname}
This research was partly supported by the Institute of Information \& Communications Technology Planning \& Evaluation (IITP)-ITRC (Information Technology Research Center) grant funded by the Korea government (MSIT) (IITP-2025-RS-2020-II201787, contribution rate: 50\%) and (IITP-2025-RS-2023-00259991, contribution rate: 50\%).

\subsubsection{\discintname}
The authors have no competing interests to declare relevant to this article’s content.
\end{credits}

%
%
%
\bibliographystyle{splncs04}
\bibliography{reference}

\begin{thebibliography}{10}
\providecommand{\url}[1]{\texttt{#1}}
\providecommand{\urlprefix}{URL }
\providecommand{\doi}[1]{https://doi.org/#1}

\bibitem{abouelmehdi2018big}
Abouelmehdi, K., Beni-Hessane, A., Khaloufi, H.: Big healthcare data: preserving security and privacy. J. Big Data  \textbf{5}(1),  1--18 (Jan 2018)

\bibitem{acarfederated}
Acar, D.A.E., Zhao, Y., Matas, R., Mattina, M., Whatmough, P., Saligrama, V.: Federated learning based on dynamic regularization. In: Proc. ICLR. Vienna, Austria (May 2021)

\bibitem{acevedo2020dataset}
Acevedo, A., Merino, A., Alf{\'e}rez, S., Molina, {\'A}., Bold{\'u}, L., Rodellar, J.: A dataset of microscopic peripheral blood cell images for development of automatic recognition systems. Data in brief  \textbf{30} (Apr 2020)

\bibitem{antunes2022federated}
Antunes, R.S., Andr{\'e}~da Costa, C., K{\"u}derle, A., Yari, I.A., Eskofier, B.: Federated learning for healthcare: Systematic review and architecture proposal. ACM Trans. Intell. Syst. Technol.  \textbf{13}(4),  1--23 (May 2022)

\bibitem{bisio2025ai}
Bisio, I., Fallani, C., Garibotto, C., Haleem, H., Lavagetto, F., Hamedani, M., Schenone, A., Sciarrone, A., Zerbino, M.: Ai-enabled internet of medical things: Architectural framework and case studies. IEEE Internet Things Mag.  \textbf{8}(2),  121--128 (Feb 2025)

\bibitem{10269141}
Dai, R., Yang, X., Sun, Y., Shen, L., Tian, X., Wang, M., Zhang, Y.: Fedgamma: Federated learning with global sharpness-aware minimization. IEEE Trans. Neural Netw. Learn. Syst.  \textbf{35}(12),  17479--17492 (Dec 2024)

\bibitem{ding2022federated}
Ding, J., Tramel, E., Sahu, A.K., Wu, S., Avestimehr, S., Zhang, T.: Federated learning challenges and opportunities: An outlook. In: Proc. IEEE ICASSP. Marina Bay, Singapore (May 2022)

\bibitem{guan2024federated}
Guan, H., Yap, P.T., Bozoki, A., Liu, M.: Federated learning for medical image analysis: A survey. Pattern Recognit.  (Jul 2024)

\bibitem{houssein2023boosted}
Houssein, E.H., Sayed, A.: Boosted federated learning based on improved particle swarm optimization for healthcare iot devices. Computers in Biology and Medicine  \textbf{163} (Sep 2023)

\bibitem{joshi2022federated}
Joshi, M., Pal, A., Sankarasubbu, M.: Federated learning for healthcare domain-pipeline, applications and challenges. ACM Trans. Comput. Healthc.  \textbf{3}(4),  1--36 (Nov 2022)

\bibitem{kairouz2021advances}
Kairouz, P., McMahan, H.: Advances and Open Problems in Federated Learning, Found. Trends Mach. Learn., vol.~14. Now Publishers (2021)

\bibitem{kather2019predicting}
Kather, J.N., Krisam, J., Charoentong, P., Luedde, T., Herpel, E., Weis, C.A., Gaiser, T., Marx, A., Valous, N.A., Ferber, D., et~al.: Predicting survival from colorectal cancer histology slides using deep learning: A retrospective multicenter study. PLoS medicine  \textbf{16}(1) (Jan 2019)

\bibitem{lee2025revisit}
Lee, Y., Gong, J., Choi, S., Kang, J.: Revisit the stability of vanilla federated learning under diverse conditions. In: Proc. MICCAI. Daejeon, Republic of Korea (Sep 2025)

\bibitem{lee2022accelerated}
Lee, Y., Park, S., Ahn, J.H., Kang, J.: Accelerated federated learning via greedy aggregation. IEEE Commun. Lett.  \textbf{26}(12),  2919--2923 (Dec 2022)

\bibitem{lee2023fast}
Lee, Y., Park, S., Kang, J.: Fast-convergent federated learning via cyclic aggregation. In: Proc. IEEE ICIP. Kuala Lumpur, Malaysia (Oct 2023)

\bibitem{li2022federated}
Li, Q., Diao, Y., Chen, Q., He, B.: Federated learning on non-iid data silos: An experimental study. In: Proc. IEEE ICDE. Kuala Lumpur, Malaysia (May 2022)

\bibitem{li2020federated}
Li, T., Sahu, A.K., Talwalkar, A., Smith, V.: Federated learning: Challenges, methods, and future directions. IEEE Signal Process. Mag.  \textbf{37}(3),  50--60 (May 2020)

\bibitem{li2019convergence}
Li, X., Huang, K., Yang, W., Wang, S., Zhang, Z.: On the convergence of fedavg on non-iid data. In: Proc. ICLR. Virtual Event, Ethiopia (May 2020)

\bibitem{mcmahan2017communication}
McMahan, B., Moore, E., Ramage, D., Hampson, S., y~Arcas, B.A.: Communication-efficient learning of deep networks from decentralized data. In: Proc. AISTAT. Fort Lauderdale, United States (Apr 2017)

\bibitem{nguyen2022federated}
Nguyen, D.C., Pham, Q.V., Pathirana, P.N., Ding, M., Seneviratne, A., Lin, Z., Dobre, O., Hwang, W.J.: Federated learning for smart healthcare: A survey. ACM Comput. Surv.  \textbf{55}(3),  1--37 (Feb 2022)

\bibitem{pfitzner2021federated}
Pfitzner, B., Steckhan, N., Arnrich, B.: Federated learning in a medical context: a systematic literature review. ACM Trans. Internet Technol.  \textbf{21}(2),  1--31 (Jun 2021)

\bibitem{qu2022generalized}
Qu, Z., Li, X., Duan, R., Liu, Y., Tang, B., Lu, Z.: Generalized federated learning via sharpness aware minimization. In: Proc. ICML. Baltimore, United States (Jul 2022)

\bibitem{rajpurkar2022ai}
Rajpurkar, P., Chen, E., Banerjee, O., Topol, E.J.: Ai in health and medicine. Nature medicine  \textbf{28}(1),  31--38 (Jan 2022)

\bibitem{rauniyar2023federated}
Rauniyar, A., Hagos, D.H., Jha, D., H{\aa}keg{\aa}rd, J.E., Bagci, U., Rawat, D.B., Vlassov, V.: Federated learning for medical applications: A taxonomy, current trends, challenges, and future research directions. IEEE Internet Things J.  \textbf{11}(5),  7374--7398 (Mar 2024)

\bibitem{rieke2020future}
Rieke, N., Hancox, J., Li, W., Milletari, F., Roth, H.R., Albarqouni, S., Bakas, S., Galtier, M.N., Landman, B.A., Maier-Hein, K., et~al.: The future of digital health with federated learning. NPJ Digit. Med.  \textbf{3}(1), ~119 (Sep 2020)

\bibitem{smith2024amd}
Smith, A., Loh, G.H., Wuu, J., Naffziger, S., Huang, T., McIntyre, H., Mangaser, R., Jung, W., Swaminathan, R.: Amd instinct™ mi300x accelerator: Packaging and architecture co-optimization. In: IEEE Symp. VLSI Circuits. Hawaii, United States (Jun 2024)

\bibitem{sun2023dynamic}
Sun, Y., Shen, L., Chen, S., Ding, L., Tao, D.: Dynamic regularized sharpness aware minimization in federated learning: Approaching global consistency and smooth landscape. In: Proc. ICML. Hawaii, United States (June 2023)

\bibitem{sunfedspeed}
Sun, Y., Shen, L., Huang, T., Ding, L., Tao, D.: Fedspeed: Larger local interval, less communication round, and higher generalization accuracy. In: Proc. ICLR. Kigali, Rwanda (May 2023)

\bibitem{whang2023data}
Whang, S.E., Roh, Y., Song, H., Lee, J.G.: Data collection and quality challenges in deep learning: A data-centric ai perspective. The VLDB Journal  \textbf{32}(4),  791--813 (Jan 2023)

\bibitem{woo2023convnext}
Woo, S., Debnath, S., Hu, R., Chen, X., Liu, Z., Kweon, I.S., Xie, S.: Convnext v2: Co-designing and scaling convnets with masked autoencoders. In: Proc. IEEE/CVF CVPR. Vancouver, Canada (Jun 2023)

\end{thebibliography}

\end{document}